\documentclass[conference,12pt,onecolumn]{IEEEtran}
\IEEEoverridecommandlockouts
\usepackage{mathptmx} 

\usepackage{cite}
\usepackage{amsmath,amssymb,amsfonts}
\usepackage{algorithmic}
\usepackage{graphicx}
\usepackage{textcomp}
\usepackage{xcolor}

\usepackage{mathtools} 
\usepackage{amsmath,amssymb,amsfonts}
\usepackage{booktabs} 
\usepackage{tikz} 

\usepackage{hyperref}
\usepackage{url}
\usepackage{xspace}
\usepackage{paralist}
\usepackage[ruled,vlined,linesnumbered]{algorithm2e}
\usepackage{subcaption}
\usepackage{mathtools}
\usepackage{textcomp}

\newcommand\pbesShort{\xspace{ }PBES\xspace}
\newcommand\datasetOne{\xspace{ }Sports73\xspace}
\newcommand\datasetTwo{\xspace{ }Sports100\xspace}
\newcommand\datasetThree{\xspace{ }Tiny ImageNet\xspace}
\newcommand\datasetFour{\xspace{ }CIFAR100\xspace}

\newcommand{\algolab}[1]{\label{algo:#1}}
\newcommand{\algoref}[1]{Algorithm~\ref{algo:#1}}

\newcommand{\ceil}[1]{\left \lceil #1 \right \rceil}

\newtheorem{lemma}{Lemma}

\newcommand{\lemlab}[1]{\label{lemma:#1}}

\newcommand{\set}[1]{\{ #1 \}}
\newcommand{\remove}[1]{}

\def\BibTeX{{\rm B\kern-.05em{\sc i\kern-.025em b}\kern-.08em
    T\kern-.1667em\lower.7ex\hbox{E}\kern-.125emX}}
\begin{document}

\title{PBES: PCA Based Exemplar Sampling Algorithm for Continual Learning \\
}
\author{\IEEEauthorblockN{Sahil Nokhwal}
\IEEEauthorblockA{\textit{Dept. Computer Science} \\
\textit{University of Memphis}\\
Memphis, USA  \\
nokhwal.official@gmail.com}
\and
\IEEEauthorblockN{Nirman Kumar}
\IEEEauthorblockA{\textit{Dept. Computer Science} \\
\textit{University of Memphis}\\
Memphis, USA \\
nkumar8@memphis.edu}}
\maketitle


\thispagestyle{plain}
\pagestyle{plain}

\begin{abstract}
We propose a novel exemplar selection approach based on Principal Component Analysis (PCA) and median sampling, and a neural network training regime in the setting of class-incremental learning. This approach avoids the pitfalls due to outliers in the data and is both simple to implement and use across various incremental machine learning models. It also has independent usage as a sampling algorithm. We achieve better performance compared to state-of-the-art methods.
\end{abstract}


\section{INTRODUCTION}
\label{sec:intro}

In \emph{continual learning} (CL) a machine learning model continually keeps learning from new data and the data is viewed as a \emph{stream} rather than a batch. A model in a CL system has to adapt to the new incoming data, and suffers from so-called \emph{catastrophic forgetting} (CF) due to the inaccessibility of the data of earlier tasks. Since a trained model depends heavily on the class distribution seen in the data, the reason for CF is that the class distributions of incoming data change over time. Newer classes are introduced with time and this invariably changes the class distribution. Conceptually the stream of data can be thought of as a sequence of \emph{tasks} where each task usually consists of data about a few new classes. Thus, different tasks have different class distributions and the model needs to update itself in order to recognize the new classes. It is impractical to retrain a model over all the past data, and thus one ``remembers'' (stores) either just a sample of the older data or else one tries to make the model itself more complex so that it can hope to just incorporate the newer data into the additional model complexity. In either case, there is some forgetting involved as the entire old data cannot be stored or remembered in the model.

The challenges posed by CF cause difficulties for neural network models to adapt to practical systems across a multitude of fields \cite{prabhu2020gdumb}.  While there have been CL studies on various domains such as image classification on commonly found day-to-day objects (CIFAR10, CIFAR100, ImageNet datasets), food images (Food1K dataset), and others, but datasets with high diversity (large inter-class gaps) have not been studied previously. Such data usually have outliers, and it becomes crucial to remember data that is not an outlier. An example domain that has this large data variance is the sports domain. In the sports domain, there is high diversity and a large inter-class gap, and thus existing approaches do not perform well while classifying images of various sports. 
Another difficulty associated with learning systems is handling class imbalance. Labeled data for classes may be highly imbalanced and this affects the model trained so that its performance can be bad. This problem is perhaps
exacerbated in a CL scenario where only a sample of the older data can be remembered. In case the data is highly imbalanced, it is possible that the rare classes get no representation in the sample at all, while
the dense classes occupy all of the samples. This can potentially lead to overfitting or underfitting of the trained model.
The goal of this paper is to address these problems. First, we want to design a more robust sampling scheme so that the data remembered is less prone to outliers. 
Second, our learning approach should be able to handle class imbalance gracefully.
There have been intriguing findings in recent rehearsal-based CL papers -- approaches that maintain a fraction of previously seen data when training new incoming classes \cite{prabhu2020gdumb, nokhwal2023dss} in class-incremental CL scenario, hence, mitigating CF. However, as mentioned, in rehearsal-based CL approaches an important question arises: how should the representative memory be managed optimally? Due to the limited number of stored data points compared to the number of incoming data points, during training the stored data points could either suffer from over-fitting or be disregarded because of the large quantity of incoming data points. A naive approach would be to progressively raise the storage size as new tasks are coming; however, this technique neglects an important representative memory constraint, i.e., to store a fixed number of data points. Hence, an approach is desired that preserves enough information about previous classes while using a modest number of data points. 

The literature \cite{rebuffi2017icarl, he2020incremental} mainly uses a herding algorithm \cite{welling2009herding} for choosing data points, also known as \emph{exemplars}, that is based only on the class mean. As per \cite{javed2018revisiting}, the herding algorithm is no better than a random selection of data points. Many researchers have proposed other effective algorithms to select exemplars in rehearsal-based methods to mitigate the CF \cite{chen2021principal,  nokhwal2023pbes}. As mentioned before however, none of the current approaches performed well in our experiments when the data variance is very large, such as in the sports domain. We propose a novel sampling algorithm that performs better than state-of-the-art methods in CL. 
Our proposed CL system is effective for both class-balanced and class-imbalanced datasets. The proposed system is effective even when the dataset is sparse and intra-class variation is high. 
To test the performance of our system in a class-imbalanced scenario, we use it for the image classification problem in the sports domain. For our experiments, we have used\datasetTwo,\datasetThree and\datasetFour datasets.
Our main contributions are as follows:
\begin{inparaenum}[(1)]
    \item A novel sampling algorithm\pbesShort to select exemplars is proposed that is robust even when outliers are present.
    \item We show how to mitigate class imbalance issues in CL settings by using KeepAugment \cite{gong2021keepaugment} -- a data augmentation approach.
    \item We demonstrate the effectiveness of our proposed method using a class-imbalanced\datasetTwo image dataset and two balanced, namely\datasetThree and\datasetFour, datasets. We demonstrate that our overall CL system outperforms existing state-of-the-art approaches in all cases.
\end{inparaenum}

\section{RELATED WORK}
\label{sec:related_work}
In this section, we mention some related work and how the work in this paper fits in the context. During an inference phase, the task ID is not present in the class-incremental (CIL) scenario, but it is present in the task-incremental (TIL) scenario. The paper focuses on CIL, which is more difficult than TIL. Even though there is no task ID, we may conceptually divide the data stream into chunks where each chunk has data for a fixed number of classes. Each such chunk can be termed as a task.
Our work is based on a rehearsal-based approach where representative memory can be used to retain some exemplars of previous tasks, which can subsequently be replayed in the current task \cite{nokhwal2023dss, rebuffi2017icarl}. 

\textbf{Representative Memory Management:} Here memory management refers to managing the \emph{content} of the memory, i.e., which exemplars to store. Various techniques are found in the literature \cite{rebuffi2017icarl,nokhwal2023dss} for this purpose. Intriguingly, considering their computational complexity, a number of suggestions \cite{rebuffi2017icarl,chaudhry2018riemannian} demonstrate only minor accuracy improvements above regular random sampling. Discriminative sampling~\cite{liu2020mnemonics}, herding selection~\cite{welling2009herding}, and samples based on entropy~\cite{chaudhry2018riemannian} are a few examples of these approaches, and even more recent work on online coreset selection is proposed in~\cite{yoon2021online}. Herding selection selects samples proportionate to the number of data points in each class, depicting each data point's distance from its class mean. Discriminative sampling selects data points that establish decision boundaries between classes. In entropy-based sampling, data points are selected based on the randomness of their softmax distributions in the final layer. Each class distribution's mean and boundary may be accurately represented by~\cite{liu2020mnemonics} suggested sampling approach. Using a cardinality-constrained bilevel optimization approach, a representative memory technique generation of a coreset is given by \cite{borsos2020coresets}. A GAN-based memory is suggested by \cite{cong2020gan} that can be perturbed for exemplars in incremental learning (IL). These papers address the effectiveness of the data points kept in memory; nonetheless, training a sample GAN for exemplars' memory would be computationally intensive or complex \cite{borsos2020coresets}. GAN-based sample generation is unstable, hence we employ effective exemplar sampling.

\section{NOTATION, FORMAL SETUP, AND BASIC ARCHITECTURE} 
\label{sec:prob_definition}
\newcommand{\classSet}{\mathcal{C}}
\newcommand{\dataSet}{\mathcal{D}}
\newcommand{\class}{c}
During CIL training, a stream of data is presented to the learning system and this can be divided into $T$ tasks. Conceptually, each task is a pair $t=(\classSet,\dataSet)$ of a set of classes $\classSet$ and data about them
$\dataSet$, so overall the stream looks like,
$(\classSet^1,\dataSet^1), (\classSet^2,\dataSet^2), \ldots, (\classSet^T,\dataSet^T)$, where $t_i=(\classSet^i,\dataSet^i)$ is the $i^{th}$ task.
We let $\classSet^i = \{ \class^i_1, \class^i_2, \ldots, \class^i_{m_i} \}$ where 
the $\class^i_j$ are the $m_i$ classes in the set $\classSet^i$. The dataset $\dataSet^i$ is the collection of points, $\{(x^i_1,y^i_1), (x^i_2,y^i_2), \ldots, (x^i_{n_i}, y^i_{n_i}) \}$ where $x^i_j$ is a data point and $y^i_j \in \classSet^i$ is its label. Thus, the total number of data points is, $\sum_{i=1}^T n_i$ and the total number of classes is $\sum_{i=1}^T m_i$. 
Notice that $\classSet^i \cap \classSet^j = \emptyset$ for $i \neq j$.

Here we will assume for CIL that the number of classes $m_i$ does not change across the tasks so $m_i$ is the same for all $i$, however, the total number of data points can vary from task to task, so $n_i$ may not equal $n_j$ for $i \neq j$. Moreover, the number of data points in $\dataSet^i$ that have $y^i_j = \class^i_k$, i.e., are of the $k^{th}$ class, may not be the same for all $k$ between $1$ and $m_i$. This is termed class imbalance.

We now describe briefly the architecture of most offline learning systems, including for example iCaRL \cite{rebuffi2017icarl}.
The main components are (1) \emph{Representative memory} which is used to store and manage data points that best represent what the model has learned in previous tasks. Storing all prior data points may be impracticable due to storage issues. To alleviate CF in class-incremental settings, rehearsal-based methods retain a small amount of previously seen data termed  \emph{exemplars}. Two important aspects of representative memory are \emph{selection and discarding techniques} which refers to which examples are stored as exemplars and which of them are discarded, 
and the \emph{memory size} which refers to how many exemplars are stored; (2) \emph{Model} - this is usually a deep neural network \cite{tanwer2020system} that has been trained on two different loss functions, such as distillation loss and cross-entropy loss, and it is typically called a cross-distilled loss function. It extracts a fixed number of features from every data point; 
(3) \emph{Classifier} - this is an algorithm that classifies a test data point into one of the classes seen so far; (4) \emph{Data augmentation} - refers to increasing the number of data points by modifying the original dataset.
\begin{figure*}[h!]
    \centering
    \includegraphics[height=6.5cm]{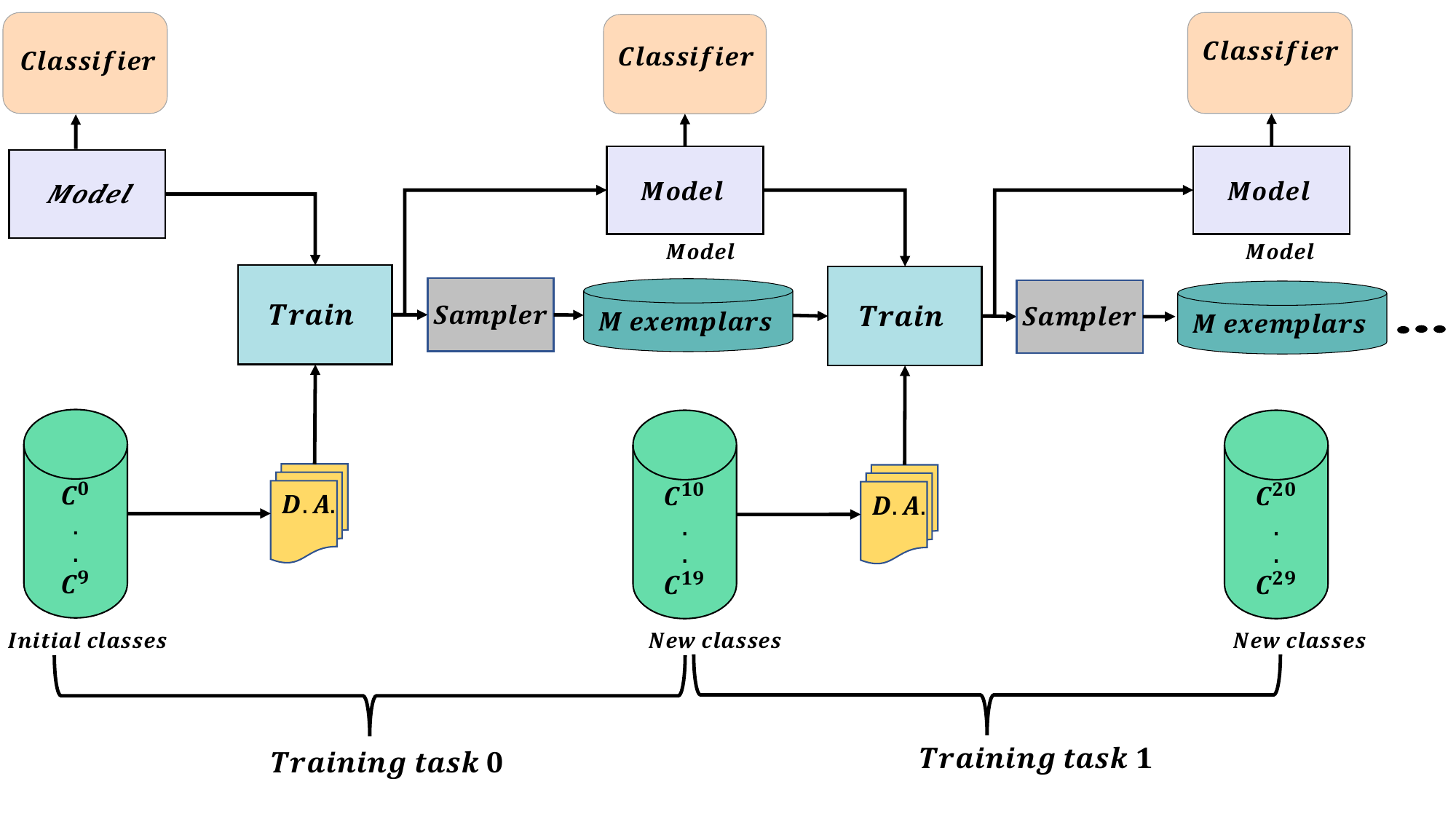}
    \caption{Architecture of Our Continual Learning Approach}
    \label{fig:our_cl_arch}
\end{figure*}

\section{OUR APPROACH}
\label{sec:our_approach}
Our proposed method to mitigate CF includes a novel technique of selecting exemplars. This technique is coupled with an effective data augmentation technique to produce a balanced training dataset from an imbalanced dataset. The overall workflow is shown in Figure~\ref{fig:our_cl_arch}. Here D.A. stands for Data Augmentation. Rather than choosing exemplars according to the class mean, as used in herding~\cite{welling2009herding}, we develop our custom sampler which is based on PCA and median sampling. The KeepAugment~\cite{gong2021keepaugment} algorithm is used to supplement each new incoming class, creating a balanced data stream. We apply the classification loss to this balanced stream and the distillation loss to the exemplars stored. In the next section, we will discuss each component’s specifics.

\subsection{Selection of exemplars}
\label{sec:exemplars_selection}
In this section, we present our algorithm,\pbesShort, to select the exemplars. The pseudocode for the algorithm is shown in \algoref{sampling}. The input to the algorithm is a training set \cite{nokhwal2023rtra} $X = \{ x_1, x_2, \ldots, x_n \}$ of $n$ images of a particular class $c$ and a target integer $m \leq n$ of images to select as exemplars. Henceforth we simply call the images ``points'' as they are treated as points in a high-dimensional space. The algorithm starts by computing 
$p$ as $\ceil{m/2}$ or $\ceil{m/2} + 1$, the latter choice being taken only when $n$ is odd and $m$ is even. 
Then, it runs the PCA algorithm on $X$ and computes the first $p$ principal directions. 
Next, it initializes a set $R$ to $X$. The set $R$ is the set of remaining points, none of which have been selected. The sequence of selected points is stored in $P$, of which the first $m$ will be returned at the end. It repeatedly
does the following: projects and sorts the remaining set $R$ of points on the $i^{th}$ principal direction (for $i=1,2,\ldots, $) and select either $1$ (when $|R|$ is odd) or
$2$ (when $|R|$ is even) median points, and appends them to $P$, removing them from $R$.

Another idea, that we use as a control, is to project in a few random directions. The idea of reducing the dimensionality
of a data set using random projections goes back to the Johnson-Lindenstrauss lemma~\cite{jl-paper}, which says that the pairwise distances between $n$ points are approximately preserved upto $1 \pm \varepsilon$, when they are projected to a random $O(\log n/\varepsilon^2)$ dimensional subspace. This projection can be achieved essentially by projecting to those many randomly chosen directions. In general, random projections are an important tool to preserve geometry and random directions serve as a good control set of directions. We implement the idea of using median sampling along a few random directions; see method RandP in Section~\ref{s:exp_results}.

The following elementary lemma gives an important property of~\algoref{sampling}.
\begin{lemma}
\lemlab{samplemma}%
The sampling algorithm appends exactly $m$ or $m + 1$ points to $P$ (before returning its first $m$ points), depending on whether $m, n$ have the same or different parity.
\end{lemma}

\subsection{Generating balanced dataset using data augmentation and training regime}
Since predicting\datasetTwo dataset class distribution is a real challenge, the model finds it hard to retain knowledge learned so far. The problem is exacerbated as the dataset is class imbalanced. The most popular continual learning research focuses on experiments with balanced datasets such as MNIST, CIFAR100, ImageNet, etc. In those datasets, the number of data points is equal between classes, whereas the number of data points in\datasetTwo is not equal. Therefore, we have proposed to use KeepAugmentation~\cite{gong2021keepaugment} algorithm to make each task $t$ of\datasetTwo balanced during the training of the model. Although a substantial amount of work has been devoted to enhancing the variety of exemplars kept in memory through data augmentation, but less work has been devoted to improving the robustness of the model using the technique. 
Data augmentation improves the classifier's generalization performance to achieve the highest efficiency on learned tasks. First, from a task $t_i=(\classSet^i,\dataSet^i)$, identify the class $c^i_j \in \classSet^i$ that has the highest number of images among all the classes in $\classSet^i$. Let $|c^i_j|$ denote its size. Then, apply the KeepAugmentation algorithm on images of all other classes in the task $t_i$ except for class $c^i_j$, and generate the required number of images $r^i_k$ to make a class $c^i_k$ balanced with respect to the class $c^i_j$, where $k \in [1, m_i], k \neq j$.
Therefore, the required number of additional augmented data points to balance the classes can be expressed as $r^i_k = |c^i_j|-|c^i_k|, k \in [1, m_i], k\neq j$. Let $I^i_k$ denote the $r^i_k$ new images generated for class $c^i_k$ and $\dataSet^i_k$ represent the original training dataset of class $c^i_k$. Then, the augmented dataset for class $c^i_k$ is $I^i_k \cup \dataSet^i_k$. The augmented dataset for each class $c^i_k$ for $k \in [1, m_i], k \neq j$, and $c^i_j$ are then used to train the model.

To balance an imbalanced dataset using the \cite{gong2021keepaugment} technique, consider an image from original training data from task $t_i$ of class $c_k^i$. The saliency map is represented as $w_{pq}{(x, y)}$,
where $x$ is the image and $y$ is its label. Here $(p, q)$ represents the pixel position of the image. The importance score for a region $S$ of the image 
is formulated as
\remove{
\begin{equation}\label{eq:sScore}
    \mathcal{I}(S, x^i_{j}, y^i_{j}) = 
    \sum_{r = 1}^{n_i}
    \sum_{(pq)\in{S}}w_{pq}(x^i_{j}, y^i_{j})
\end{equation}
}
$\mathcal{I}(S, x^i_{j}, y^i_{j}) =     
    \sum_{(p,q)\in{S}}w_{pq}(x^i_{j}, y^i_{j})$.
Using a thresholding approach we identify a region with low importance score. This region is then cut out to get the augmented image and this approach is known as the selective cut method. The augmented image denoted
as $\tilde{x}^i_j$ can be computed as, $
    \tilde{x}^i_j = (1-M(S)) \odot x^i_j,
$
where $\odot$ refers to pointwise multiplication and $M(S)$ is the binary mask defined by,
$M(S) = \left[M_{pq}(S)\right]_{pq}$, where $M_{pq} = \mathbb{I}((p, q)\in{S})$.

\begin{algorithm}[h]
    \DontPrintSemicolon
    \SetKwFunction{FSampler}{PBES\_Sampler}
    \SetKwProg{Fn}{Function}{:}{}
    \Fn{\FSampler{$X, m$}}{
        \KwIn{Image training set $X = \{x_1, \mathellipsis, x_n\}$ of class $c$, $m$ -- number of exemplars to output}
        \KwOut{$P$ exemplars}

        \BlankLine

        \If{$n$ is odd and $m$ is even}{
            $p = \ceil{m/2} + 1$\;
        }
        \Else{
            $p = \ceil{m/2}$\;
        }
        
        Compute the $p$ principal directions of $X$ using PCA\;
        $R \gets X$\;
        \For{$i \gets 1$ to $p$}{
            $L_i =$ Sort points of $R$ along $i^{th}$ principal direction \;
            \If{$|R|$ is even}{
                Let $x_1, x_2$ be the lower and higher median points from $L_i$\;
                $P \gets P + \langle x_1, x_2 \rangle$\;
                $R \gets R \setminus \{ x_1, x_2 \}$\;
            }
            \Else{
                Let $x_1$ be the median point from $L_i$\;
                $P \gets P + \langle x_1 \rangle$\;
                $R \gets R \setminus \{ x_1 \}$\;
            }
        }
        \Return first $m$ points of $P$
    }
\caption{Median points sampling of exemplars}
\algolab{sampling}%
\end{algorithm}

\subsection{Loss equations formulation for the model}
Let $\dataSet_{M}^i$ represent the training data at task $t_i$ - this is fed to train the neural network for the current model. $\dataSet^i$ depicts the original data from the input stream, $I^i$ is the augmented data generated from the original data stream and $\dataSet_{E}$ represents the stored exemplars in rehearsal memory.

We have that, 
$\dataSet_{M}^i = \dataSet^i \cup I^i \cup \dataSet_{E}$. For a given input $\textbf{x}$, the current model's output logits are represented as
\begin{equation}
    p^{(|C_{(i-1)}| + m_i)} (\dataSet_M^i(\textbf{x})) = \\\left( L^{(1)},\cdots, L^{(|C_{(i-1)}|)}, L^{(|C_{(i-1)}|+1)}, \cdots, L^{(|C_{(i-1)}|+m_i)} \right),
\end{equation}
while the output logits of the teacher model are
\begin{equation}
     \hat{p}^{(|C_{(i-1)}|)}(I_E(\textbf{x})) = 
    \left( \hat{L}^{(1)},\cdots, \hat{L}^{(|C_{(i-1)}|)} \right)
\end{equation}

Therefore, the cross-entropy loss for training task $t_i$ is formulated as
\begin{equation}\label{eq:crossLoss}
    \mathcal{L}_C{(\dataSet_{M}^i}) = \sum_{\textbf{x} \in \dataSet_{M}^i} \sum_{j = 1}^{|C_{(i-1)}|+m_i} -\hat{y}^{(j)} \log[p^{(j)}_T(\dataSet_{M}^i(\textbf{x}))],
\end{equation}
where $|C_{i-1}|$ represents the total number of classes already experienced by the neural network model until task $t_{i-1}$, $m_i$ denotes the number of new classes that came in the current task $t_i$, $\hat{y}$ represents the label data in one-hot encoding format for the input $x$, and the probabilities $p^j_T$ are defined below. 
The knowledge distillation loss for retaining the information from the older model is,
\begin{equation}\label{eq:kdLoss}
    \mathcal{L}_\dataSet{(\dataSet_{M}^i, I_{E})} 
    = \sum_{\textbf{x} \in \dataSet_{M}^i} \sum_{j = 1}^{|C_{i-1}|} -\hat{p}_{T}^{(j)}(I_{E}(\textbf{x})) \log[p_{T}^{(j)}(\dataSet_{M}^i(\textbf{x}))],
\end{equation}
where
\begin{align*}
\hat{p}_T^{(j)}(I_E(\textbf{x})) &= \dfrac{\exp(\hat{L}^{(j)}/T)}{\sum_{d=1}^{|C_{i-1}|}\exp(\hat{L}^{(d)}/T)}, \\
p_T^{(j)}(\dataSet_M^i(\textbf{x})) &= \dfrac{\exp(L^{(j)}/T)}{\sum_{d=1}^{|C_{i-1}|}\exp(L^{(d)}/T)}.
\end{align*}
Here a temperature scalar value $T > 1$ is used to regulate the disparity between two distributions. Therefore, the final loss of the model can be formulated as
\begin{equation}
    \mathcal{L}_{CD}(\dataSet_{M}^i) \\
    = \beta \mathcal{L}_D{(\dataSet_{M}^i,I_E)} + (1-\beta) \mathcal{L}_C{(\dataSet_{M}^i)}.
\end{equation}
Here $0 \leq \beta \leq 1$; for this work the value of $\beta$ is 0.5.

\section{Experimental results}
\label{s:exp_results}%
In our experiments, we first compare our approach with the well-known state-of-the-art approaches such as Rainbow Memory\cite{bang2021rainbow}, GDUMB\cite{prabhu2020gdumb}, and iCaRL\cite{rebuffi2017icarl}. Further, we compare our approach to two other approaches, fine-tuning and upper-bound. As part of the fine-tune approach, only the new class images are used for training, and exemplars are ignored. Cross-entropy loss is applied to the new class images, and distillation loss is not taken into consideration. Thus, the number of exemplars is zero. The fine-tune approach is taken as a lower bound for our problem (in terms of performance). In the lower bound, the exemplars are not stored in memory, and the training for all the tasks happens without considering any exemplars. The upper-bound approach considers all the images seen so far at every task and applies cross-entropy loss. In short, the number of exemplars per class is equal to the number of images for that particular class. We have also compared our results with random projections. Section~\ref{s:ablation_study} of this paper discusses the effectiveness of our exemplar selection through ablation studies.

\begin{figure*}[h!]
    \centering
    \includegraphics[height=\textwidth]{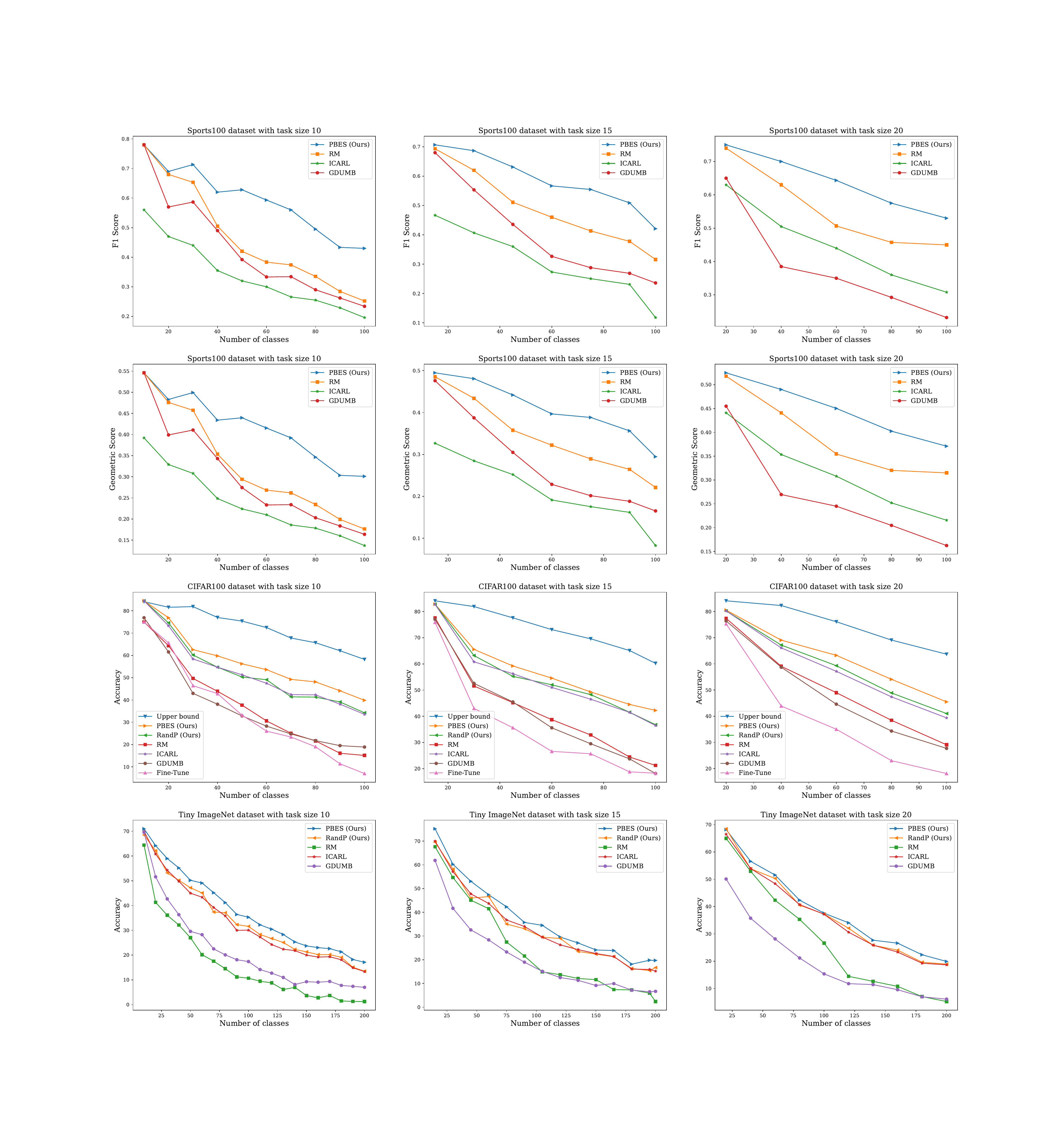}
    \caption{Row 1 displays the F1 score of\datasetTwo with task sizes 10, 15, and 20. Row 2 displays the Geometric mean score of\datasetTwo with task sizes 10, 15, and 20. Row 3 displays the Accuracy of\datasetFour with task sizes 10, 15, and 20. Row 4 displays the Accuracy of\datasetThree with task sizes 10, 15, and 20.}
    \label{fig:results}
\end{figure*}

\subsection{Comparative study with other existing techniques}
     We observe that the accuracy of the CL approach changes depending on the task size. For a certain number of training classes to learn, a lower task size will result in more incremental tasks, increasing the likelihood of catastrophic forgetting. In contrast, training many classes for every incremental task is likewise a tough challenge for bigger task sizes. In particular, we discover that fine-tune suffers from a severe catastrophic forgetting problem, with Last and Average accuracy being drastically worse than they are with the upper-bound. This is because there is limited training data for every learned task throughout the IL process. In Figure~\ref{fig:results}, rows 1 and 2, display the F1 and Gmean scores, respectively, of\datasetTwo examined at the end of every task with task sizes of 10, 15, and 20. Our approach achieves better results than state-of-the-art methods in continual learning. The performance difference between our approach and the upper-bound is minimal. The results for\datasetFour and\datasetThree datasets considering only proposed\pbesShort sampler without KeepAugmentation technique are shown in figure~\ref{fig:results}, rows 3 and 4, respectively. 
   
    
    
\subsection{Ablation study}%
\label{s:ablation_study}%
    Now, we will study our approach's components and illustrate their influence on the final accuracy\cite{nokhwal2023embau}. The number of exemplars is kept fixed for these studies. Our approach has two modules 1) module-1: our proposed algorithm\pbesShort, and, 2) module-2: our approach to integrate the KeepAugment algorithm \cite{gong2021keepaugment} for data augmentation. In particular, we compare and contrast the following approaches: 
    \begin{inparaenum}[(1)]
        \item \textbf{baseline:} select exemplars using the herding algorithm.
        \item \textbf{\pbesShort sampler without KeepAugmentation :}\pbesShort sampler for exemplar selection but no data augmentation.
        \item \textbf{baseline and KeepAugmentation method:} Herding algorithm for exemplar selection and KeepAugmentation for data augmentation.
        \item \textbf{Ours:}\pbesShort sampler for exemplar selection and  KeepAugmentation for data augmentation.
    \end{inparaenum}
    Each component (\pbesShort and KeepAugmentation) discussed in this paper, when compared to the baseline, improves performance. We find that when we integrate all the modules of the recommended strategy, our results significantly outperform the baseline. Since there is a significant class imbalance present in the\datasetTwo dataset, where the number of data points in training data varies in the range [98, 191] among sports classes, we also conclude that our training regime gives significant performance improvements for such dataset. The dataset also has huge variance and the proposed strategy performs well even in this case. 
    
    \begin{table}[h!]
    \label{tbl:ablation_study}
    \centering
    \caption{Average accuracy of different methods on\datasetTwo dataset with task size 20}
    \begin{tabular}{*{3}{|c}|}
    \hline
        \bfseries Method & \bfseries \datasetTwo \\ \hline
        baseline & 44.54 \\ \hline
        \pbesShort w/o KeepAugment & 63.56 \\ \hline
        baseline w/ KeepAugment  & 61.28 \\ \hline
        \pbesShort w/ KeepAugment & 68.27 \\ \hline
      \end{tabular}
    \end{table}

\subsection{Influence of Exemplars Size}%
\label{s:infl_ex_sz}%
    This section shows the effect of exemplars size on the performance. To test this we vary the total number of stored exemplars $M$ in $\set{800, 1000, 1200, 1400}$ for\datasetTwo. Our average accuracy using the\datasetTwo dataset with a task size of 20 is displayed in Table~\ref{tbl:exemplars_effect_tbl2}. For both approaches, improved performance is seen when more exemplars are used. Exemplar memory capacity is amongst the most crucial components for CL systems, especially in an offline environment. We discover that our proposed technique performs better than the baseline for any given number of exemplars. This improvement over baseline becomes more apparent as the number of allowed exemplars decreases, i.e., our method responds much better under a memory crunch.

    

\begin{table}[ht]
\centering
\caption{Influence of Exemplars Size for\datasetTwo dataset}
  \begin{tabular}{ |c|c|c|c| } 
    \hline
    $M$=800 & $M$=1000 & $M$=1200 & $M$=1400 \\ \hline
    63.97 & 66.4  & 69.2 & 72.23  \\ \hline
  \end{tabular}
  \label{tbl:exemplars_effect_tbl2}
\end{table}

\section*{Conclusions}
This paper proposes a novel sampling method that selects representative data from incoming training data to reduce catastrophic forgetting. A learning regime using data augmentation to create a balanced training set from an unbalanced dataset was also presented. The complex\datasetTwo,\datasetThree, and\datasetFour datasets showed that our strategy outperforms state-of-the-art methods.

\medskip
\bibliographystyle{ieeetr}
\bibliography{my_bib}

\clearpage

\appendix

\section{Variance comparison of \datasetTwo and \datasetFour datasets}
While comparing the \datasetTwo and \datasetFour datasets, we found that not only does the \datasetTwo dataset have a large inter-class variance as compared to the \datasetFour dataset, but its intra-class variance is also larger than the \datasetFour dataset over almost all classes.

\datasetTwo has an average intra-class variance of \textbf{[0.53702784, 0.52929306, 0.5373449]}, where three numbers represent the red, green, and blue channels of a dataset. For \datasetFour, the average intra-class variance is \textbf{[0.51339644, 0.5058607, 0.5224528]}. We have added variances for 10 randomly chosen classes of\datasetFour and\datasetTwo datasets.

\textbf{\datasetTwo} \\
Class:3 Variance:  [0.532834,  0.5222238,  0.52689004] \\
Class:4 Variance:  [0.5306238,  0.51647824,  0.5180705] \\
Class:7 Variance:  [0.53316474,  0.5186189,  0.51601034] \\
Class:12 Variance:  [0.53707814,  0.52304155,  0.52288795] \\
Class:19 Variance:  [0.5373644,  0.52640015,  0.52934366] \\
Class:23 Variance:  [0.5345275,  0.52474713,  0.52843] \\
Class:48 Variance:  [0.53620374,  0.53087264,  0.5425352 ] \\
Class:50 Variance:  [0.53640944,  0.53035456,  0.54162234] \\
Class:60 Variance:  [0.53711337,  0.53135484,  0.5414798] \\
Class:91 Variance:  [0.53706926,  0.5309249,  0.5437807]

\textbf{\datasetFour} \\
Class:3, Variance:  [0.5230189,  0.5120296,  0.51592296] \\
Class:4, Variance:  [0.51551026,  0.5054853,  0.5091805] \\
Class:7, Variance:  [0.51268476,  0.5044046,  0.51462716] \\
Class:12, Variance:  [0.5126156,  0.5079753,  0.5226499] \\
Class:20, Variance:  [0.51169854,  0.5073559,  0.5250322] \\
Class:23, Variance:  [0.5122692,  0.50743765,  0.5242459] \\
Class:48, Variance:  [0.51306504,  0.5067717,  0.52280056] \\
Class:50, Variance:  [0.5126162,  0.50611514,  0.52281636] \\
Class:60, Variance:  [0.5129891,  0.5049722,  0.52449954] \\
Class:91, Variance:  [0.511761,  0.5018764,  0.5216153] \\
Class:91, Variance:  [0.511761,  0.5018764,  0.5216153] \\

\newpage
\section{Histogram to visualize class imbalance in \datasetOne dataset}
A histogram of \datasetOne dataset for random 25 classes is illustrated which shows the imbalance of different classes in the dataset. The total number of classes ranges between 99 and 191. 
\begin{figure}[h!]
    \centering    \includegraphics[width=0.8\textwidth,height=0.85\textheight]{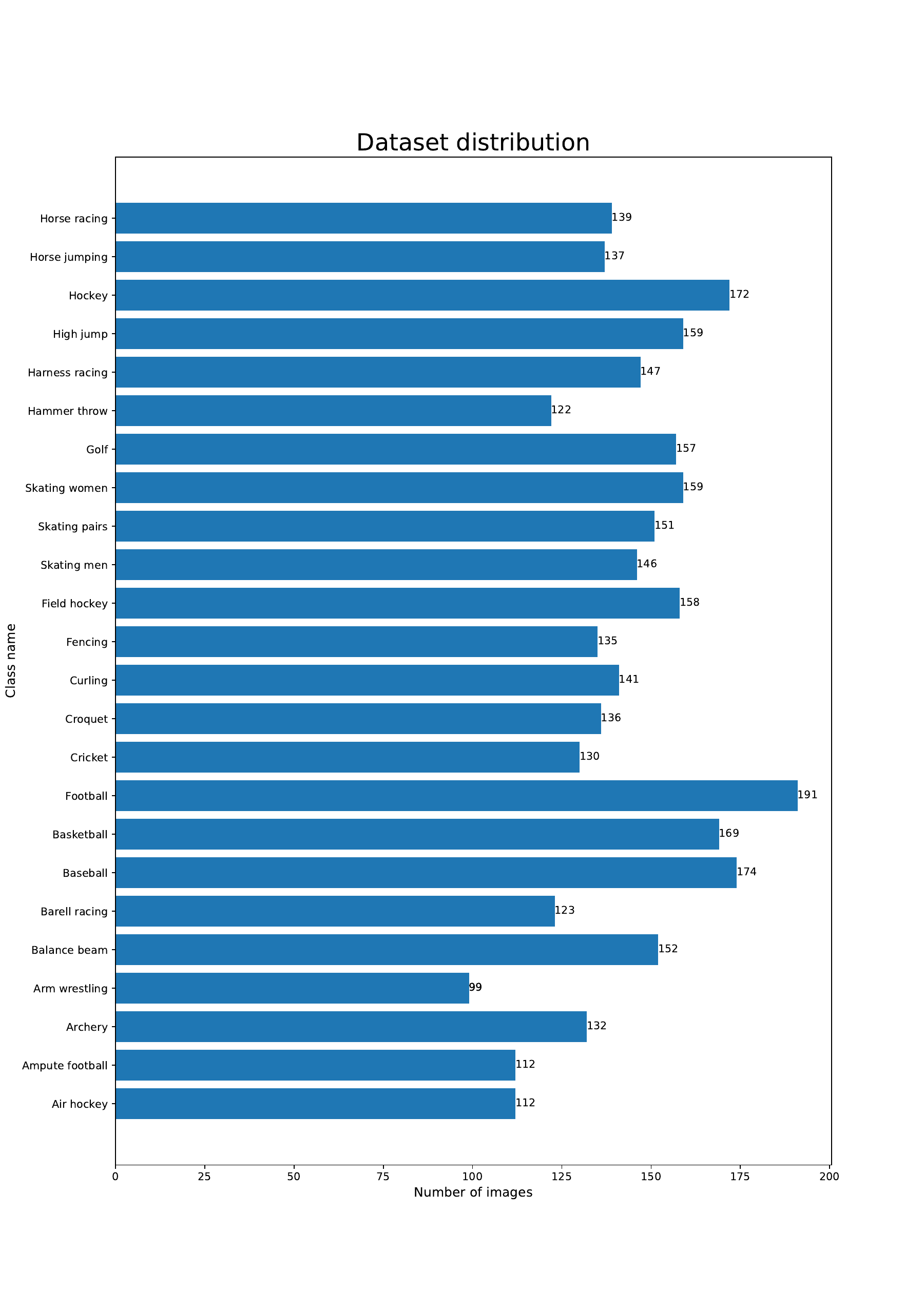}
    \caption{Histogram of \datasetOne classes}
    \label{fig:sports73Histogram}
\end{figure}




\end{document}